\documentclass[11pt, a4paper]{article}
\usepackage{latexsym}
\usepackage[fleqn]{amsmath}
\usepackage{graphicx}
\usepackage[fleqn]{amsmath}
\usepackage[varg]{txfonts}
\usepackage[english]{babel}
\usepackage[hyperindex=false,
    pageanchor=false,
    pdfmenubar=false,
    pdfpagelabels=false, breaklinks=true]{hyperref}
\usepackage{natbib}
\bibpunct{(}{)}{,}{a}{,}{,}

\usepackage{setspace}
\usepackage{epstopdf}
\usepackage{setspace}
\usepackage{graphicx}
\usepackage{threeparttable}

\newcommand{\mb}{\mathbb} 
\newcommand{\vc}{\mathbf} 

\newcommand{\mc}{\mathcal}
\newcommand{\mr}{\mathrm}

\begin{document}
\title{Identifying Outliers   using Influence Function of  Multiple Kernel Canonical Correlation Analysis} 
\author{\textbf{  Md. Ashad Alam$^{1,2}$,  and Yu-Ping Wang$^1$}  \\ $^{1}$Department of Biomedical Engineering, Tulane University\\
 New Orleans, LA 70118, USA\\
$^2$Department of Statistics, Hajee Mohammad Danesh Science and Technology\\ University, Dinajpur 5200, Bangladesh}       
\date{}
\maketitle
\begin{abstract}
Imaging genetic research has essentially focused on discovering unique and co-association effects, but typically ignoring to identify outliers or atypical objects in genetic as well as non-genetics variables. Identifying  significant outliers is an essential and challenging issue for imaging genetics and multiple sources data analysis. Therefore, we need to examine for transcription errors of identified outliers. First, we address the influence function (IF) of kernel mean element, kernel covariance operator, kernel cross-covariance operator, kernel canonical  correlation analysis (kernel CCA) and multiple kernel CCA. Second, we propose an  IF of multiple kernel CCA, which can be applied for more than two datasets. Third, we propose a visualization method to detect influential observations of multiple sources of data based on the IF  of kernel CCA and multiple kernel CCA. Finally, the proposed methods are capable of analyzing outliers of subjects usually found in biomedical applications, in which the number of dimension is large. To examine the outliers, we use the stem-and-leaf display. Experiments on both synthesized and imaging genetics data (e.g., SNP, fMRI, and DNA methylation) demonstrate that the proposed visualization can be applied effectively.
\end{abstract}  

\section{Introduction}
The problem of identifying significant outliers is an essential and challenging issue in statistical machine learning for multiple sources data analysis. The atypical objects or  outliers, data  that cause surprise in relation to the majority of the data, often occur in the real data. Outliers may be right, but we need to examine for transcription errors. They can play havoc with classical statistical methods \citep{Gogoi-11}. Once a statistical approach is applied to imaging genetics data containing outliers, the results can be deceptive with high probability. To overcome this problem, since $1960$ many robust methods have been developed, which are less sensitive to outliers. The goals of robust statistics are to use the methods from the bulk of the data and identify the points deviating from the original patterns for further investment \citep{Huber-09,Hampel-11,Nasser-12}. But it is well-known that most robust methods are computationally intensive and have the curse of dimensionality problem. The outliers need to be removed or downweighted prior to fitting non-robust statistical or machine learning approaches \citep{Filzmoser-08,Oh-08,Roth-06}.

The incorporation of various unsupervised learning methods into genomic analysis is a rather recent topic. Using the dual representations, the task of learning from multiple data sources is related to the kernel-based data integration, which has been actively studied in the last decade \citep{Hofmann-08,Ashad-14T}. Kernel fusion in unsupervised learning has a close connection with unsupervised kernel methods. As unsupervised kernel methods, kernel principal component analysis \citep{Schlkof-kpca,Ashad-14}, kernel canonical correlation analysis \citep{Akaho,Ashad-15,Ashad-13}, weighted multiple kernel CCA have been extensively studied for decades \citep{Yu-11}. But these methods are not robust; they are sensitive to contaminated data.  To apply all of these non-robust methods, for instance in genomics, outliers identification and/ or robust approaches are essential.

Due to the properties of eigen decomposition, kernel CCA is still a well-applied method for multiple sources data analysis and integration. An empirical comparison and sensitivity analysis for robust linear CCA and kernel CCA were also discussed, which give similar interpretation as kernel PCA without any theoretical results \citep{Ashad-10,Ashad-08}. In addition, \citep{Romanazii-92} and \citep{Ashad-16} have proposed the IF of canonical correlation and kernel CCA but the IF of multiple kernel CCA has not been studied. All of these considerations motivate us to conduct studies on the IF of  multiple kernel CCA to identify outliers in imaging genetics data sets: SNP, fMRI, and DNA methylation.

The contribution of this paper is fourfold. We address the IF of kernel mean element (kernel ME), kernel covariance operator (kernel CO), kernel cross-covariance operator (kernel  CCO),  kernel canonical correlation analysis (kernel CCA) and multiple kernel CCA. After that, we propose the IF of multiple kernel CCA, which can be applied for more than two datasets.  Based on this results, we propose  a visualization method  to detect influential observations of multiple sources  data based. The proposed method is capable of analyzing the outliers usually found in biomedical application, in which the number of dimension is large. To confirm the outliers, we use the  step-and-leaf display. The results imply that the proposed method enables to identify outliers in  synthesized and imaging genetics data (e.g., SNP, fMRI, and DNA methylation).

The remainder of the paper is organized as follows. In the next section, we provide a brief review of  kernel ME, kernel CO, and kernel  CCO. In Section $3$,  we discuss in  brief  the IF, IF of kernel ME and IF of kernel CO. After a brief review of kernel CCA in Section \ref{sec:CKCCA}, we propose the IF of classical multiple kernel CCA in Section \ref{sec:MKCCA}. In Section $5$, we describe experiments conducted on both synthesized and real data analysis from an imaging genetics study with a visualizing method.

\section{Kernel Mean element and kernel covariance operator}
Kernel ME, kernel CO and kernel CCO with positive definite kernel have been extensively applied to nonparametric statistical inference through representing distribution in the form of means and covariance in RKHS \citep{Gretton-08, Fukumizu-08,Song-08, Kim-12,Gretton-12}. Basic notations of   kernel MEs, kernel CO and kernel CCO with their robustness through IF are briefly discussed below.

\subsection{Kernel mean element}
\label{sec:kernel ME}
Let $F_X$, $F_Y$  and $F_{XY}$ be the probability measure on $\mc{X}$, $\mc{Y}$ and  $ \mc{X}\times\mc{Y}$, respectively.
Also let $X_1, X_2, \ldots, X_n$,;  $ Y_1, Y_2, \ldots, Y_n$ and $(X_1, Y_1),(X_2, Y_2), \ldots, (X_2, Y_2)$ be the random sample from the respective distribution. A symmetric kernel $k(\cdot,\cdot)$ defined on a space is called {\em positive definite kernel} if the Gram matrix $(k(X_i, X_j))_{ij}$ is positive semi-definite \citep{Aron-RKHS}. By the reproduction properties and kernel trick, the kernel can evaluate the inner product of any two feature vectors efficiently without knowing an explicit form of either the {\em feature map} ($\Phi(\cdot) = k(\cdot, X), \forall X\in \mc{X}$) or {\em feature space} ($\mc{H}$). In addition,  the computational cost does not depend on the dimension of the original space after computing the Gram matrices \citep{Fukumizu-14,Ashad-14}.
A mapping $\mc{M}_X:= \mb{E}_X[\Phi(X)] = \mb{E}_X[k(\cdot, X)]$ with $\mb{E}_X[\sqrt{k(X, X)}] < \infty$ is an element of the RKHS $\mc{H}_X$. By the reproducing property with $X\in \mc{X}$, {\em kernel mean element} is defined as \[\langle \mc{M}_X, f \rangle_{\mc{H}_X} = \langle \mb{E}_X[k(\cdot, X)], f \rangle_{\mc{H}_X} = \mb{E}_X[f(X)],\] for all    $f\in \mc{H}_X$. Given an  independent and identically distributed sample, the mapping $m_X=\frac{1}{n}\sum_{i=1}^n\Phi(X_i)=  \frac{1}{n} \sum_{i=1}^n k(\cdot, X_i)$ is an empirical element of the RKHS, $\mc{H}_X$,
$\langle m_X, f \rangle_{\mc{H}_X}= \langle  \frac{1}{n}\sum_{i=1}^n k(\cdot, X_i), f\rangle =  \frac{1}{n}\sum_{i=1}^n  f(X_i).
$

\subsection{Kernel covariance operator}
\label{sec:kernel CO}
By the reproducing property, kernel CCO, $\Sigma_{XY} := \mc{H}_Y \to \mc{H}_X$ with $\mb{E}_X[k_X(X, X)] < \infty$, and $\mb{E}_Y[k_Y(Y, Y)] < \infty$ is defined as\begin{eqnarray}
\langle f_X,  \Sigma_{XY}f_Y\rangle_{\mc{H}_X}&=& \mb{E}_{XY}\left[\langle f_X, k_X(\cdot, X) - \mc{M}_X \rangle_{\mc{H}_X} \langle f_Y, k_Y(\cdot, Y) - \mc{M}_Y \rangle_{\mc{H}_Y} \right]\nonumber\\&=&\mb{E}_{XY}\left[(f_X(X) - E_X[f(X)]) (f_Y(Y) - E_Y[f(Y)])\right]. \nonumber
\end{eqnarray}
 \section{Influence function of kernel operators}
To define the notation of robustness in statistics, different approaches have been proposed, for examples, the {\em minimax approach} \citep{Huber-64}, the {\em sensitivity curve} \citep{Tukey-77}, the {\em influence functions} \citep{Hampel-74,Hampel-86} and in the finite sample {\em breakdown point} \citep{Dono-83}. Due to its simplicity, the IF is the most useful approach in statistical supervised learning \citep{Christmann-07,Christmann-04}. In this section, we briefly discuss the notations of IF, IF of kernel ME, and IF of kernel CO and kernel CCO.

Let ($\Omega$, $\mc{A}$) be a probability space and $(\mc{X}, \mc{B})$ a measure space. We want to estimate the parameter $\theta \in \Theta$  of a distribution $F$ in $\mc{A}$. We assume that there exists a functional $R: \mc{D}(R) \to \mb{R}$, where $ \mc{D}(R)$ is the set of all probability distribution in $\mc{A}$.
Let $G$ be a distribution in $\mc{A}$. If data do not fallow the model $F$ exactly but slightly  going toward $G$, the  G\^{a}teaux Derivative at $F$ is called {\em influence function} \citep{Kim-12}. The IF of complicated statistics, which is a function of simple statistics, can be calculated with the chain rule, Say
$R(F)=a(R_1(F), ....., R_s(F))$. Specifically,
 \begin{eqnarray}
IF_R(z)=\sum_{i=1}^s \frac{\partial a}{\partial R_i}IF_{R_i }(z). \nonumber
 \end{eqnarray}
It can also be used to find the IF of a transformed statistic, given the influence function for the statistic itself.

The IF of kernel CCO, $R(F_{XY})$, with joint distribution, $F_{XY}$, using complicated statistics at  $Z^\prime=(X^\prime,Y^\prime)$ is  denote as $\rm{IF}(\cdot, Z^\prime, R, F_{XY})$ and given by
\begin{align*}
\langle k_X(\cdot, X^\prime)-\mc{M}[F_X], f \rangle_{\mc{H}_X} \langle k_Y(\cdot,Y^\prime)\mc{M}[F_Y], g \rangle_{\mc{H}_Y}
-\mb{E}_{XY}[ \langle k_X(\cdot, X)-\mc{M}[F_X], f \rangle_{\mc{H}_X} \langle k_Y(\cdot,Y)-\mc{M}[F_Y], g \rangle_{\mc{H}_Y}],
\end{align*}
which is estimated with the data points $(X_1 Y_1), (X_2, Y_2)$,$ \cdots,$ $(X_n, Y_n) \in \mc{X}\times \mc{Y}$ for every $Z_i = (X_i, Y_i)$ as
\begin{multline}
\widehat{\rm{IF}}( Z_i, Z^\prime, R, F_{XY})  \\
=  [k_X(X_i, X^\prime)-\frac{1}{n}\sum_{b=1}^n k_X(X_i, X_b)] [k_Y(Y_i, Y^\prime)\\-\frac{1}{n}\sum_{b=1}^n k_Y(Y_i, Y_b)] -
\frac{1}{n}\sum_{d=1}^n[k_X(X_i, X_d)-\frac{1}{n}\sum_{b=1}^n k_X(X_i, X_b)]  [k_Y(Y_i, Y_d)-\frac{1}{n}\sum_{b=1}^n k_Y(Y_i, Y_b)]. \nonumber
\end{multline}
For the bounded kernels, the above IFs have three properties: gross error sensitivity, local shift sensitivity and rejection point. These are not true for the unbounded kernels, for example, liner and polynomial kernels. We are able to make similar conclusion for the kernel CO and kernel CCO. Most of the unsupervised methods explicitly or implicitly depend on the kernel CO or kernel CCO. They are sensitive to contaminated data, even when using the bounded positive definite kernels. To overcome the problem, the outliers need to removed from the data. 
\section{ Kernel CCA and multiple kernel CCA}
In this section, we review the kernel CCA, the IF and empirical IF (EIF) of kernel CCA. After that we  address the multiple kernel CCA and proposed the IF and EIF of multiple kernel CCA based on the IF of  kernel CO and kernel CCO.
\subsection{Kernel CCA}
\label{sec:CKCCA}
The aim of {\em  kernel CCA} is to seek two sets of functions in the RKHS for which the correlation (Corr) of random variables is maximized. Given two sets of random variables $X$ and $Y$ with two   functions in the RKHS, $f_{X}(\cdot)\in \mc{H}_X$  and  $f_{Y}(\cdot)\in \mc{H}_Y$, the optimization problem of the random variables $f_X(X)$ and $f_Y(Y)$ is
\begin{eqnarray}
\label{ckcca1}
\max_{\substack{f_{X}\in \mc{H}_X,f_{Y}\in \mc{H}_Y \\ f_{X}\ne 0,\,f_{Y}\ne 0}}\mr{Corr}(f_X(X),f_Y(Y)).
\end{eqnarray}
The optimizing functions $f_{X}(\cdot)$ and $f_{Y}(\cdotp)$ are determined up to scale.

Using a finite sample, we are able to estimate the desired functions. Given an i.i.d sample, $(X_i,Y_i)_{i=1}^n$ from a joint distribution $F_{XY}$, by taking the inner products with elements or ``parameters" in the RKHS, we have features
$f_X(\cdot)=\langle f_X, \Phi_X(X)\rangle_{\mc{H}_X}= \sum_{i=1}^na_X^ik_X(\cdot,X_i) $ and
 $f_Y(\cdot)=\langle f_Y, \phi_Y(Y)\rangle_{\mc{H}_Y}=\sum_{i=1}^na_Y^ik_Y(\cdot,Y_i)$, where $k_X(\cdot, X)$ and $k_Y(\cdot, Y)$ are the associated kernel functions for $\mc{H}_X$ and $\mc{H}_Y$, respectively. The kernel Gram matrices are defined as   $\vc{K}_X:=(k_X(X_i,X_j))_{i,j=1}^n $ and $\vc{K}_Y:=(k_Y(Y_i,Y_j))_{i,j=1}^n $.  We need the centered kernel Gram matrices $\vc{M}_X=\vc{C}\vc{K}_X\vc{C}$ and $\vc{M}_Y=\vc{C}\vc{K}_Y\vc{C}$, where $\vc{C} = \vc{I}_n -\frac{1}{n}\vc{B}_n$ with $\vc{B}_n = \vc{1}_n\vc{1}^T_n$ and $\vc{1}_n$ is the vector with $n$ ones. The empirical estimate of Eq. (\ref{ckcca1}) is then given by
\begin{eqnarray}
\label{ckcca6}
\max_{\substack{f_{X}\in \mc{H}_X,f_{Y}\in \mc{H}_Y \\ f_{X}\ne 0,\,f_{Y}\ne 0}}\frac{\widehat{\rm{Cov}}(f_X(X),f_Y(Y))}{[\widehat{\rm{Var}}(f_X(X))]^{1/2}[\widehat{\rm{Var}}(f_Y(Y))]^{1/2}} \nonumber
\end{eqnarray}
where
\begin{align*}
& \widehat{\rm{Cov}}(f_X(X),f_Y(Y))
= \frac{1}{n} \vc{a}_X^T\vc{M}_X\vc{M}_Y \vc{a}_Y \\
& \widehat{\rm{Var}}( f_X(X))
=\frac{1}{n} \vc{a}_X^T\vc{M}_X^2 \vc{a}_X \,  \\ &\widehat{\rm{Var}}( f_Y(Y))=\frac{1}{n} \vc{a}_Y^T\vc{M}_Y^2 \vc{a}_Y,
\end{align*}
where $\vc{a}_{X}$ and $\vc{a}_{Y}$ are the directions of $X$ and $Y$, respectively. After using simple algebra, We can write
\begin{align}
\label{ckcca7}
\begin{bmatrix}
   0      & \vc{M}_1\vc{M}_2  \\
\vc{M}_2\vc{M}_1      & 0  \\
 \end{bmatrix}
\begin{bmatrix} \vc{a}_{X}\\
                \vc{a}_{Y}\\
  \end{bmatrix} =  \rho\begin{bmatrix}
   \vc{M}_1\vc{M}_1     & 0  \\
0     &   \vc{M}_2\vc{M}_2  \\
\end{bmatrix}
\begin{bmatrix} \vc{a}_{X}\\
               \vc{a}_{Y}\\
  \end{bmatrix}  \end{align}
Unfortunately, the naive kernelization (\ref{ckcca7}) of CCA  is trivial and non-zero solutions of generalized eigenvalue problem are $\rho=\pm 1$ \citep{Ashad-10, Back-02}. To overcome this problem, we introduce small regularization terms in the denominator of the right hand side of (\ref{ckcca7}) as
\begin{align}
\label{ckcca8}
\begin{bmatrix}
   0      & \vc{M}_1\vc{M}_2  \\
\vc{M}_2\vc{M}_1      & 0  \\
 \end{bmatrix}
\begin{bmatrix} \vc{a}_{X}\\
                \vc{a}_{Y}\\
  \end{bmatrix} = \rho\begin{bmatrix}
   (\vc{M}_1+\kappa I)^2     & 0  \\
0     &   (\vc{M}_2+\kappa I)^2  \\
\end{bmatrix}
\begin{bmatrix} \vc{a}_{X}\\
               \vc{a}_{Y}\\
  \end{bmatrix}  \end{align}
where the small regularized coefficient is $\kappa >0$.

Using the IF of kernel mean element and covariance operator in the eigenvalue problem in Eq. (\ref{ckcca8}), as shown in \citep{Ashad-16} , the influence function of kernel canonical correlation (kernel CCA ) and kernel canonical variate at $Z^\prime = (X^\prime, Y^\prime)$ is given by
\begin{multline}
\label{TIFKCCA}
\rm{IF} (Z^\prime, \rho_j^2)= - \rho_j^2 \bar{f}_{jX}^2(X^\prime) + 2 \rho_j \bar{f}_{jX}(X^\prime) \bar{f}_{jY}(Y^\prime)  - \rho_j^2 \bar{f}_{jY}^2(Y^\prime),\\
\rm{IF} (\cdot, Z^\prime, f_{jX}) = -\rho_j (\bar{f}_{jY}(Y^\prime) - \rho_j \bar{f}_{jX}(X^\prime))\mb{L} \tilde{k} (\cdot, X^\prime) -  (\bar{f}_{jX}(X^\prime)  - \rho_j \bar{f}_{jY}(Y^\prime))\mb{L} \Sigma_{XY}\Sigma^{-1}_{YY} \tilde{k}_Y(\cdot,  Y^\prime)+\frac{1}{2}[1- \bar{f}^2_{jX}(X^\prime)]f_{jX},
\end{multline}
\rm{where} $\mb{L}= \Sigma_{XX}^{- \frac{1}{2}}(\Sigma_{XX}^{- \frac{1}{2}} \Sigma_{XY} \Sigma_{YY}^{-1} \Sigma_{YX}\Sigma_{XX}^{- \frac{1}{2}}-\rho^2\vc{I})^{-1}\Sigma_{XX}^{- \frac{1}{2}}$ and similar for the kernel CV of $f_Y$, $\rm{IF} (\cdot, Z^\prime, f_{jY})$
It is known that the inverse of an operator may not exit or even exist but may not be continuous in general \citep{Fukumizu-SCKCCA}. While we can derive kernel canonical correlation using correlation operator
$ \vc{V}_{YX}=  \Sigma_{YY}^{- \frac{1}{2}}\Sigma_{YX} \Sigma_{XX}^{- \frac{1}{2}}$, even when $\Sigma_{XX}^{- \frac{1}{2}}$ and  $\Sigma_{YY}^{- \frac{1}{2}}$ are not proper operators, the IF of covariance operator is true only for the finite dimensional RKHSs. For infinite dimensional RKHSs,  we can find IF of $\Sigma_{XX}^{- \frac{1}{2}}$  by introducing  a regularization term as follows
\begin{multline}
\rm{IF}(\cdot, X^\prime,
(\Sigma_{XX} + \kappa\vc{I})^{- \frac{1}{2}}) = \frac{1}{2} [(\Sigma_{XX}+\kappa\vc{I})^{- \frac{1}{2}}- (\Sigma_{XX}+\kappa\vc{I})^{- \frac{1}{2}}\tilde{k}_X(\cdot, X^\prime)\otimes  \tilde{k}_X(\cdot, X^\prime)(\Sigma_{XX} + \kappa\vc{I})^{- \frac{1}{2}}], \nonumber
\end{multline}

where $\kappa > 0$ is a regularization coefficient, which gives an empirical estimator. Let $(X_i, Y_i)_{i=1}^n$ be a sample from the distribution $F_{XY}$. The EIF of Eq.(\ref{TIFKCCA}) at $Z^\prime=(X^\prime, Y^\prime)$ for all points $Z_i = (X_i, Y_i)$ are
\begin{align}
\label{EIFKCCA}
 &\rm{EIF} (Z^\prime , \rho_j^2)= \widehat{\rm{IF}} (Z^\prime,  \hat{\rho}_j^2), \nonumber\\
&\rm{EIF} (Z_i, Z^\prime, f_{jX}) = \widehat{\rm{IF}} (\cdot, Z^\prime, f_{jX}), \nonumber\\
 &\rm{EIF} (Z_i, Z^\prime, f_{jY})= \widehat{\rm{EIF}} (\cdot, Z, \widehat{f}_{jY}),
\end{align}
respectively.

For the bounded kernels the IFs or EIFs, which are stated in Eq.(\ref{TIFKCCA}), they have the three properties: gross error sensitivity, local shift sensitivity and rejection point. But for unbounded kernels, say  a linear or polynomial, the IFs are not bounded. As a consequence, the results of classical kernel CCA using the bounded kernels are less sensitive than classical kernel CCA using the unbounded kernels \citep{Ashad-16, Huang-KPCA}.

\subsection{Multiple kernel CCA}
\label{sec:MKCCA}
Multiple  kernel CCA  seeks  more than  two sets of  functions in the RKHSs for which the correlation (Corr) of  random variables  is maximized. Given $p$ sets of random variables $X_1,  \cdots X_p$   and $p$ functions in the RKHS, $f_{1}(\cdot)\in \mc{H}_1$,$\cdots$, $f_{p}(\cdot)\in \mc{H}_p$, the optimization problem of  the random variables $f_1(X_1)$, $\cdots$, $f_p(X_p)$ is
\begin{eqnarray}
\label{mkcca1}
\max_{\substack{f_{1}\in \mc{H}_{X_i},\cdots, f_{p}\in \mc{H}_{X_i} \\ f_{1}\ne 0,\, \cdots, f_{p}\ne 0}}\sum_{j=1, j^\prime > j}^p\mr{Corr}(f_j(X_j),f_j^\prime(X_j^\prime)).
\end{eqnarray}
Given an i.i.d sample, $(X_{i1},X_{i2}, \cdots, X_{ip})_{i=1}^n$ from a joint distribution $F_{X_1, \cdots, X_p}$, by taking the inner products with elements or ``parameters" in the RKHS, we have  features
\begin{eqnarray}
\label{mkcca2}
f_1(\cdot)=\langle f_1, \Phi_1(X_1)\rangle_{\mc{H}_1}= \sum_{i=1}^na_{i1}k_1(\cdot,X_i), \nonumber\\
\vdots,\nonumber\\
 f_p(\cdot)=\langle f_p, \phi_p(X_p)\rangle_{\mc{H}_p}=\sum_{i=1}^na_{ip}k_p(\cdot,X{_ip}),
\end{eqnarray}
where $k_1(\cdot, X_1), \cdots,  k_p(\cdot, X_p)$ are the associated kernel functions for $\mc{H}_1, \cdots, \mc{H}_p$, respectively. The kernel Gram matrices are defined as   $\vc{K}_1:=(k_1(X_{i1},X_{i^\prime 1}))_{i,i^\prime=1}^n$, $\cdots$, $\vc{K}_pY:=(k_1(X_{ip},X_{i^\prime p}))_{i,i^\prime=1}^n$. Similar to  Section \ref{sec:CKCCA}, using this kernel Gram matrices, the centered kernel Gram matrices are defined as  $\vc{M}_1=\vc{C}\vc{K}_1\vc{C}$, $\cdots$,  $\vc{M}_p=\vc{C}\vc{K}_p\vc{C}$, where $ \vc{C} = \vc{I}_n -\frac{1}{n}\vc{B}_n$ with  $\vc{B}_n = \vc{1}_n\vc{1}^T_n$ and $\vc{1}_n$ is the vector with $n$ ones. As in the two sets of data the empirical estimate of Eq. (\ref{mkcca1}) is obtained using the generalized eigenvalue problem, as given by
following problem:
\begin{align}
\label{mkcca3}
\begin{bmatrix}
   0      & \vc{M}_1\vc{M}_2 &\vc{M}_1\vc{M}_3 & \dots & \vc{M}_1\vc{M}_p \\
\vc{M}_2\vc{M}_1      & 0 & \vc{M}_2\vc{M}_3 & \dots & \vc{M}_2\vc{M}_p \\
    \hdotsfor{5} \\
    \vc{M}_p\vc{M}_1     & \vc{M}_p\vc{M}_2  & \vc{M}_p\vc{M}_3  & \dots & 0
\end{bmatrix}
\begin{bmatrix} a_1\\
                a_2\\
\hdotsfor{1}\\
a_p
  \end{bmatrix} = \rho\begin{bmatrix}
   (\vc{M}_1+\kappa I)^2     & 0 &0 & \dots & 0 \\
0     &   (\vc{M}_2+\kappa I)^2  & 0& \dots & 0 \\
    \hdotsfor{5} \\
    0     & 0  & 0 & \dots &  (\vc{M}_p+\kappa I)^2
\end{bmatrix}
\begin{bmatrix} a_1\\
               a_2\\
\hdotsfor{1}\\
a_p
  \end{bmatrix}  \end{align}
Using the similar procedure as shown in \citep{Ashad-16} and in Eq. (\ref{TIFKCCA}) to  Eq. (\ref{mkcca3}), we can easily derive the IF of multiple kernel CCA.     
The IF of $l$-th multiple kernel CCA at $Z^\prime = (X^\prime_1,\cdots, X^\prime_P)$ is expressed as
\begin{eqnarray}
\label{TIFMKCCA}
\rm{IF} (Z^\prime, \rho_l^2)= - \rho_l^2 \sum_{j=1}^p\bar{f}_{lj}^2(X^\prime_j) + 2 \sum_{j=1, j^\prime >j}^p\rho_j\bar{f}_{{lj}}(X^\prime_j)  \bar{f}_{{lj^\prime}}(X^\prime_{j^\prime})
\end{eqnarray}
\section{Experiments}
\label{Sec:Exp}
We demonstrate the experiments on synthesized and real imaging genetics data analysis including SNP, fMRI, and DNA methylation. For synthesized experiments, we generate two types of  data: original data  and those with $5\%$ of contamination, which are called ideal data (ID) and  contaminated data (CD), respectively. In all experiments, for the bandwidth of Gaussian kernel we use the median of the  pairwise distance \citep{Gretton-08,Sun-07}. Since the goal is to find the outlier, the regularization parameter of kernel CCA is set as $\kappa = 10^{-5}$. The description of real data sets is in Sections \ref{Sec:MCIC} and the synthetic data sets are described as follows:

\textbf {Multivariate Gaussian structural data (MGSD):}
Given multivariate normal data, $\vc{Z}_i\in\mb{R}^{12} \sim \vc{N}(\vc{0},\Sigma)$ ($i= 1, 2, \ldots, n$) where  $\Sigma$ is the same as in \citep{Ashad-08}. We divide $\vc{Z}_i$ into two sets of variables ($\vc{Z}_{i1}$,$\vc{Z}_{i2}$), and use the first 6 variables of $\vc{Z}_i$ as $X$ and perform $\log$ transformation of the absolute value of the remaining variables ($\log_e(|\vc{Z}_{i2}|))$) as $Y$. For the CD $\vc{Z}_i\in\mb{R}^{12} \sim \vc{N}(\vc{1},\Sigma)$ ($i= 1,2,\ldots, n$).

\textbf {Sign and cosine function structural data (SCSD):}
We use uniform  marginal distribution, and transform the data by two periodic $\sin$ and $\cos$ functions to make two sets $X$ and $Y$, respectively, with additive Gaussian noise:
$Z_i\sim U[-\pi,\pi], \,\eta_i\sim N(0,10^{-2}),~\,i=1,2,\ldots, n,
 X_{ij}=\sin(jZ_i)+\eta_i,\,  Y_{ij} = \cos(jZ_i)+\eta_i, j=1,2,\ldots, 100.$
For the CD $\eta_i\sim N(1,10^{-2})$.

\textbf {SNP and fMRI structural data (SMSD):}
Two sets of SNP data X with $1000$ SNPs and fMRI data Y with 1000 voxels were simulated. To correlate the SNPs with the voxels, a  latent model is used  as in \citep{Parkhomenko-09}). For data contamination, we consider the signal level, $0.5$ and noise level, $1$ to $10$ and $20$, respectively.

In  the experiments, first, for the effect of kernel CCA  we compared ID with CD.  To measure the influence, we calculated the ratio between ID and CD of IF of kernel CC and kernel CV. Based on this ratio, we define two measures for kernel CC and kernel CV
\begin{eqnarray}
\eta_{\rho} &=&\left | 1- \frac{\|EIF(\cdot, \rho^2) ^{ID}\|_F}{\|EIF(\cdot, \rho^2)^{CD}\|_F}\right| \qquad \rm{and} \\
\eta_{f} &=& \left| 1- \frac{\|EIF(\cdot, f_X)^{ID}- EIF(\cdot,f_Y)^{ID}\|_F}{\|EIF(\cdot, f_X) ^{CD}-EIF(\cdot, f_Y)^{CD}\|_F}\right|,\nonumber
\end{eqnarray}
respectively. The method does not depend on the contaminated data, and the above measures, $\eta_{\rho}$  and $\eta_{f}$, should  be approximately zero. In other words, the best methods should give smallest values. To compare, we consider simulated data sets:  MGSD, SCSD, SMSD with 3 sample sizes, $n\in \{ 100, 500, 1000\}$. For each sample size, we repeat the experiment for $100$ samples.  Table \ref{tbl:idcd} presents the results (e.g., mean $\pm$ standard deviation) of kernel CCA. From this table, we observe that kernel CCA  is affected by the contaminated data in all cases.
\begin{table*}
 \begin{center}
\caption {The mean and standard deviation of the measures, $\eta_{\rho}$ and $\eta_{f}$ of kernel CC and kernel CV.}
\label{tbl:idcd}
 \begin{tabular}{llcccccccc} \hline
& &\multicolumn{2}{c}{\rm{Measure}}\\ \cline{3-4}
Data&n&$\eta_{\rho}$&$\eta_{f}$  \\ \hline
&$100$&$1.9114\pm 3.5945$&$ 1.3379\pm 3.5092$\tabularnewline
MGSD&$500$&$ 1.1365\pm 1.9545$&$ 0.8631 \pm 1.3324 $\tabularnewline
&$1000$&$ 1.1695\pm 1.6264$&$ 0.6193 \pm 0.7838$\\ \hline
&$100$&$0.4945\pm 0.5750$& $1.6855\pm 2.1862$\tabularnewline
SCSD &$500$&$0.2581\pm 0.2101$& $1.3933\pm 1.9546$\tabularnewline
&$1000$&$0.1537\pm 0.1272$&$1.6822\pm 2.2284$ \\\hline
&$100$&$ 0.6455 \pm 0.0532$& $ 0.6507 \pm 0.2589 $\tabularnewline
SMSD&$500$&$0.6449\pm 0.0223$&$ 3.7345 \pm 2.2394$\tabularnewline
&$1000$&$ 0.6425 \pm 0.0134$&  $ 7.7497\pm 1.2857$\\ \hline
\end{tabular}
\end{center}
\end{table*}
\begin{table*}
 \begin{center}
\caption {The mean and standard deviation of the differences between training and testing correlation of $10$ fold cross-validation kernel CCA.}
\label{tbl:SCV}
 \begin{tabular}{cccccccc} \hline
\rm{Data}&n&$\rm{ID}$&$\rm{CD}$ \\ \hline
&$500$ &$0.7005\pm 0.0744$&$0.7536\pm 0.0503$\\
\rm{MGSD}&$1000$&$0.6459\pm 0.0234$&$0.5322\pm 0.1184$\\
&$2000$&$0.4151\pm 0.210$&$0.4673\pm 0.1196$\\
&$500$&$0.3601\pm 0.3132$&$0.2974\pm 0.3433$\\
\rm{SCSD}&$1000$&$0.0005\pm 0.0.0002$&$0.0.0003\pm 0.0.0005$\\
&$2000$&$0.0002\pm 0.0.0001$&$0.0.0003\pm 0.0.0002$\\\hline
\end{tabular}
\end{center}
\end{table*}
\begin{figure*}
\begin{center}
\includegraphics[width=16cm, height=8cm]{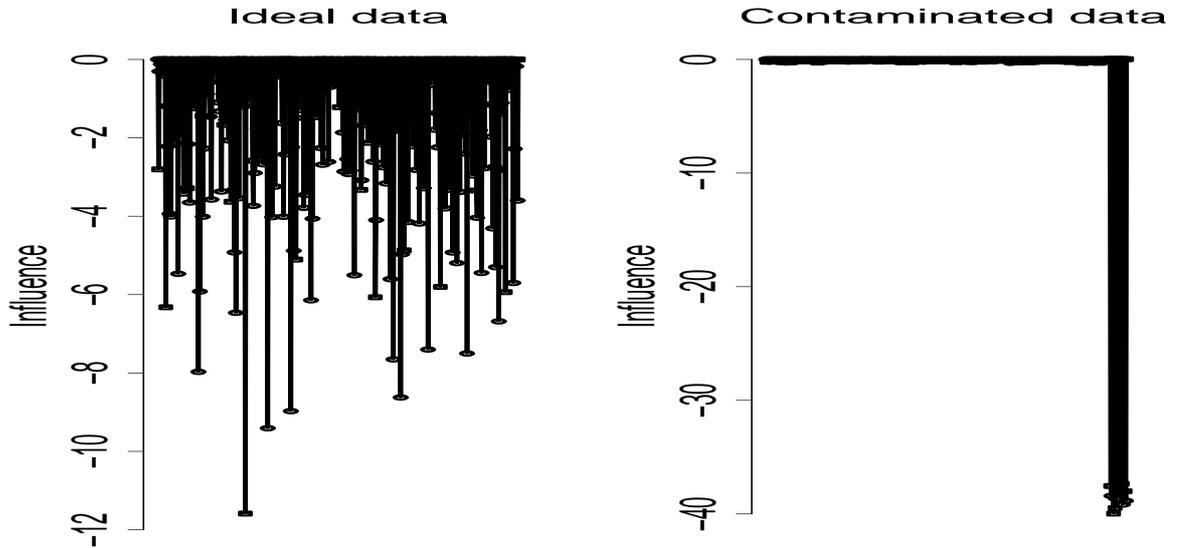}
\caption{Influence  points  using  empirical influence function of kernel CCA of of SMSD (ideal data and  contaminated data).}
\label{SMDSIFOB}
\end{center}
\end{figure*}

\begin{figure*} 
\begin{center}
\includegraphics[width=16cm, height=18cm]{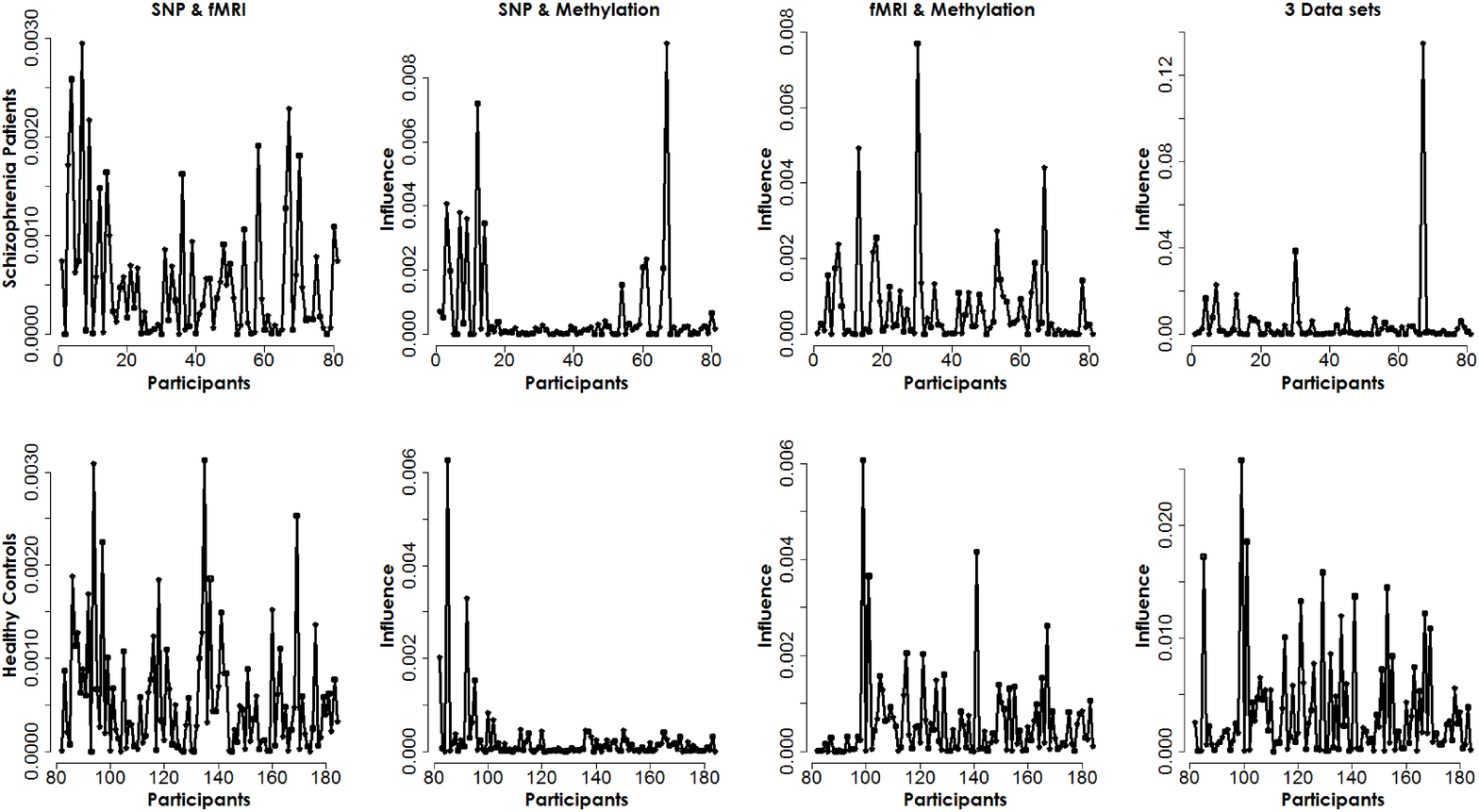}
\caption{The influence subject of MCIC data set using kernel CCA and multiple kernel CCA.}
\label{MCICFOB}
\end{center}
\end{figure*}

\begin{table*}
 \begin{center}
\caption {The stem-and-leaf display of the influence MCIC data (the decimal point is 4 digits for pairwise data and 2 digits for 3 data sets) using kernel CCA and multiple kernel CCA.}
\label{tbl:stem}
 \begin{tabular}{|cl|cl|cl|cl|} \hline
\multicolumn{2}{|c|}{\rm{SNP \& fMRI}}&\multicolumn{2}{|c|}{\rm{SNP \& Methylation}}&\multicolumn{2}{|c|}{\rm{fMRI \&Methylation}}&\multicolumn{2}{|c|}{\rm{3 Data sets}}\\ \hline
$0 |$&$ 00000000+35$&$0 |$&$00000000+77$&$0 |$&$ 00000000+46
$&$0 |$&$00000000+147$\\
$1 |$&$ 00111223+1$&$1 |$&$00012223+7$&$1 |$&$01122256
$&$1 |$&$01122345$\\
$2 |$&$01223334$&$2 |$&$01122223+2$&$2 |$&$01222333
$&$2 |$&$36$\\
$3 |$&$11234557$&$3 |$&$11112456$&$3 |$&$01113378$&$3|$&$9$\\
$4 |$&$03357788$&$4 |$&$1223557
$&$4 |$&$2333455
$&$4|$&\\
$5 |$&$01367788$&$5 |$&$2$&$5 |$&$11123567$&$\vdots$&\\
$6 |$&$00012334$&$6 |$&$158$&$6 |$&$ 1345668$&$\vdots$&\\
$7 |$&$02455779$&$7 |$&$2$&$7 |$&$ 0568 $&$\vdots$&\\
$8 |$&$446799$&$8 |$&$2$&$8 |$&$ 3334778 $&$\vdots$&\\
$9 |$&$14$&$\vdots $&$$&$9 |$&$248 $&$\vdots$&\\
$10 |$&$0016799$&$32| $&$8$&$10 |$&$036699 $&$\vdots$&\\
$\vdots $&$$&$34| $&$9$&$\vdots $&$ $&$\vdots$&\\
$21 |$&$7$&$36 |$&$0$&$36 |$&$6 $ &$$$13|$&$6$\\
$22 |$&$59$&$37 |$&$4$&$41 |$&$2$&&$$\\
$25 |$&$39$&$40 $&$8$&$44 $&$2 $&$$&\\
$29 |$&$5$&$62 |$&$6$&$49 |$&$2 $&$$&\\
$30 |$&$9$&$72 |$&$0$&$60 |$&$7 $&$$&\\
$31|$&$3$&$90 |$&$7$&$77 |$&$0 $&$$&\\ \hline
\end{tabular}
\end{center}
\end{table*}
\begin{table*}
 \begin{center}
\caption {Mean and standard deviation of  the  differences between taring and test correlation of $10$ fold cross-validation of MICI  data using kernel CCA  and multiple kernel CCA.}
\label{tbl:ifnorm}
 \begin{tabular}{llcccccccc} \hline
\rm{Data}&outliers&$\rm{All}$&$\rm{Without~ outliers}$ \\ \hline
\rm{SNP \& fMRI}& $\{ 135,94,7,4,169,67,97,9\}$ &$0.8208\pm 0.2382$&$0.7353\pm 0.1870$\\
\rm{SNP \& Methylation}& $\{ 67,12,85,3,7,9,14,92\}$ &$0.7337\pm 0.2000$&$0.6606\pm 0.1772$\\
\rm{fMRI \&Methylation}&$\{ 30,99,13,67,141,101\}$ &$0.7424\pm 0.1893$&$0.7817\pm 0.1759$\\
 \rm{SNP}, \rm{fMRI \& Methylation}&$\{67,30,99,7\}$ &$1.4115\pm 0.2544$&$0.1.2659\pm 0.1744$\\\hline
\end{tabular}
\end{center}
\end{table*}

\subsection{Visualizing influential observation using kernel CCA and multiple kernel CCA}
\label{Sec:Visu}
Now, we propose a simple graphical display based on the EIF of kernel CCA, and the index plots (the data on $x$-axis and the influence of observation, as shown in Eq. (\ref{EIFKCCA}) on $y$ axis), to assess the related influence data points in data integration with respect to EIF of kernel CCA. To do this, we first consider simulated  SMSD and then real imaging genomic  dataset (see \ref{Sec:MCIC}).  The index plots of $500$ observations using the SMSD (ID and $5\%$ CD) and influence functions based on the EIF of kernel CCA are presented in Figure \ref{SMDSIFOB}. The plots show that the influence of ID and  CD  has significance difference. On the one hand, the observations only for ID have less influence; on the other hand, the observations with CD have large influence. It is clear that the kernel CCA is affected by the CD significantly. In addition, using the visualization of the EIF of kernel CCA, we can easily identify the influence observations properly.

\subsection{Real data Analysis: Mind Clinical Imaging Consortium}
\label{Sec:MCIC}
The Mind Clinical Imaging Consortium (MCIC) has collected three types of data (SNPs, fMRI and DNA methylation) from 208 subjects including $92$ schizophrenic patients (age: $34\pm 11$, $22$ females) and $116$ (age: $32\pm 11$, $44$ females) healthy controls. Without missing data, the number of subjects is $184$ ($81$ schizophrenia (SZ) patients and $103$ healthy controls)\citep{Dongdong-14}.

{\bf SNPs}: For each subject (SZ patients  and healthy controls) a blood sample was taken and DNA was extracted. All subject genes typing was performed at the Mind Research Network using the Illumina Infinium HumanOmni1- Quad assay covering  $1140419$ SNP loci.  To form the final genotype calls and to perform a series of standard quality control procedures bead studio and PLINK software packages were applied, respectively. The final dataset spans  $722177$ loci having $22442$ genes  based on $184$ subjects (those without missing data). Genotypes   ``aa"  (non-minor allele), ``Aa" (one minor allele) and ``AA" (two minor alleles) were coded as $0$, $1$ and $2$for each  SNP, respectively \citep{Dongdong-14} \citep{Chen-12}.

{\bf fMRI}: Participants' fMRI data was collected during their block design motor response to auditory stimulation. State-of-the-art approaches use mainly Participants' feedback and experts' observations for this purpose. The aim was to continuously monitor the patients, acquiring images with parameters (TR=2000 ms, TE= 30ms, field of view=22cam, slice thickness=4mm, 1 mm skip, 27 slices, acquisition matrix $64\times 64$, flip angle =$90^{\circ}$) on a  Siemens3T Trio Scanner and 1.5 T Sonata with echo-planar imaging (EPI). Data were pre-processed with SPM5 software  and  were realigned spatially normalized and resliced to $3\times 3 \times 3$ mm.  It was  smoothed with  a  $10\times 10 \times 10$ $mm^3$ Gaussian kernel and analyzed by multiple regression considering the stimulus and their temporal derivatives plus an intercept term as repressors .  Finally the stimulus-on versus stimulus-off contrast images were extracted with $53\times 63 \times 46$ mission measurements, excluding voxels without measurements. $41236$ voxels were extracted from $116$ ROIs based on the aal brain atlas for analysis \citep{Dongdong-14}.

{\bf DNA methylation}:DNA methylation is one of the main epigenetic mechanisms to regulate gene expression. It appears to be involved in the development of SZ. In this paper, we investigated $27481$ DNA methylation markers in blood from $81$ SZ patients and $103$ healthy controls. Participants come from the MCIC, a collaborative effort of 4 research sites.  For  more details, site information and enrollment for SZ patients and healthy controls are in \citep{Liu-13}. All participants' symptoms were evaluated by the Scale of the Assessment of Positive Symptoms and the Scale of the Assessment of Negative symptoms \citep{Andreasen-84}. DNA from blood samples was measured by the Illumina Infinium Methylation27 Assay. The methylation value is calculated by taking the ratio of the methylated probe intensity and the total probe intensity.

To detect influential subjects (in SZ patients and healthy controls), as discussed in Section \ref{Sec:Visu}, we use the EIF of kernel CC of kernel CCA and multiple kernel CCA. Figure \ref{MCICFOB} shows the influence of participants from MICI data: SNPs, fMRI and DNA methylation. The  SZ patients and healthy controls are in $1$st and $2$nd rows, respectively. The analysis results of pairwise data sets (i.e., SNP \& fMRI, SNP \& Methylation, and fMRI\& Methylation) using kernel CCA and all $3$ data sets, SNP, fMRI, \& Methylation using multiple kernel CCA are in column $1$st to $4$th, respectively.  These plots show that in all scenarios the healthy controls have less influence than the SZ patients group.

To extract the outliers of subjects from participants of MCIC data, we consider stem-and-leaf display of influence of MCIC data (e.g., SNP \& fMRI, SNP \& Methylation, and fMRI\& Methylation) using kernel CCA and all $3$ data sets, SNP, fMRI, \& Methylation using multiple kernel CCA.  Table  \ref{MCICFOB} shows the results of pairwise datasets and $3$ datasets together. Based on the stem-and-leaf display, the outliers of subject sets of SNP \& fMRI, SNP \& Methylation, and fMRI\& Methylation, and SNP, fMRI, \& Methylation are $\{135,94,7,4,169,67,97,9\}$, $\{ 67,12,85,3,7,9,14,92\}$,\\  $\{30,99,13,67,141,101\}$, $\{67,30,99,7\}$. It is noted that, multiple kernel CCA is able to extract common SZ patient $67$, which is also outlier for all pairwise results using kernel CCA. Finally, we investigated the difference between training correlation and test correlation using $10$ fold cross-validation with all subjects with or without outliers. Table  \ref{tbl:ifnorm} shows the outliers of subjects along with the result of all subjects with or without outliers using kernel CCA and multiple kernel CCA. We see that after removing the outliers by the proposed methods, both kernel CCA and multiple kernel CCA performed better using all subjects.

\section{Concluding remarks and future research}
 The methods for identifying  outliers in imaging genetics data presented in this paper are not only applicable to single data sets but also for integrated data sets, which is an essential and challenging issue for multiple sources data analysis. The proposed methods are based on the IF of kernel CCA and multiple kernel CCA, which can detect and isolate the outlier effectively in both synthesized and real data sets.  After applying to pairwise data (e.g., SNP \& fMRI, SNP \& Methylation, and fMRI\& Methylation) using kernel CCA and to all $3$ data sets (e.g., SNP, fMRI, \& Methylation) using multiple kernel CCA,  we found that in all scenarios the healthy controls have less influence than the SZ patients. In addition,  multiple kernel CCA is able to extract the common  SZ patient $67$, which is also the outliers for all pairwise data analysis using kernel CCA. After removing the significant outliers indicated by both kernel CCA and multiple kernel CCA, the stem-and-leaf display shows that both methods performed much better than using all subjects.

Although we have argued that the kernel CCA and multiple kernel CCA procedure for detecting outliers worked effectively, there is also space for further improvement.  The use of the Gaussian kernel function is an optimal selection; however, other classes of kernel functions may be more reasonable  for a specific data set.  In future work, it would be also interesting to develop robust kernel PCA and robust multiple kernel CCA and apply them to imaging genomic analysis.
\subsection*{Acknowledgments}
The authors wish to thank the NIH (R01 GM109068, R01 MH104680) and NSF (1539067) for support.

\begingroup
\bibliographystyle{plainnat}
\bibliography{Ref-UKIF}
\endgroup

\end{document}